\documentclass[default,iicol]{sn-jnl}

\usepackage{array}
\usepackage{booktabs}
\usepackage{color}
\usepackage{xcolor}
\usepackage{pifont}
\usepackage{subcaption}

\def\ie{\emph{i.e.}}
\def\eg{\emph{e.g.}}
\newcommand{\etal}{\textit{et al}.}

\jyear{2021}%

\theoremstyle{thmstyleone}%
\theoremstyle{thmstyletwo}%

\theoremstyle{thmstylethree}%

\begin{document}

\title[LCVSL]{Local Compressed Video Stream Learning for Generic Event Boundary Detection}

\author[]{Libo Zhang$^{1,2,3}\dagger$}\email{libo@iscas.ac.cn}

\author[]{Xin Gu$^{2}\dagger$}\email{guxin21@mails.ucas.ac.cn}

\author[]{Congcong Li$^{2}$}\email{licongcong18@mails.ucas.edu.cn}

\author[]{Tiejian Luo$^{2}$}\email{tjluo@ucas.ac.cn}

\author*[]{Heng Fan$^{4},*$}\email{heng.fan@unt.edu}

\affil[1]{Institute of Software Chinese Academy of Sciences, Beijing, China}

\affil[2]{University of Chinese Academy of Sciences, Beijing, China}

\affil[3]{Nanjing Institute of Software Technology, Nanjing, China}

\affil[4]{Department of Computer Science and Engineering, University of North Texas, Denton, TX, USA}


\abstract{Generic event boundary detection aims to localize the generic, taxonomy-free event boundaries that segment videos into chunks. Existing methods typically require video frames to be decoded before feeding into the network, which contains significant spatio-temporal redundancy and demands considerable computational power and storage space. To remedy these issues, we propose a novel compressed video representation learning method for event boundary detection that is fully end-to-end leveraging rich information in the compressed domain, \ie, RGB, motion vectors, residuals, and the internal group of pictures (GOP) structure, without fully decoding the video. Specifically, we use lightweight ConvNets to extract features of the P-frames in the GOPs and spatial-channel 
attention module (SCAM) is designed to refine the feature representations of the P-frames based on the compressed information with bidirectional information flow. To learn a suitable representation for boundary detection, we construct the local frames bag for each candidate frame and use the long short-term memory (LSTM) module to capture temporal relationships. We then compute frame differences with group similarities in the temporal domain. This module is only applied within a local window, which is critical for event boundary detection. Finally a simple classifier is used to determine the event boundaries of video sequences based on the learned feature representation. To remedy the ambiguities of annotations and speed up the training process, we use the Gaussian kernel to preprocess the ground-truth event boundaries. Extensive experiments conducted on the Kinetics-GEBD and TAPOS datasets demonstrate that the proposed method achieves considerable improvements compared to previous end-to-end approach while running at the same speed. The code is available at \url{https://github.com/GX77/LCVSL}.}

\keywords{Generic Event Boundary Detection (GEBD), Spatial-Channel Attention Module (SCAM), Group Similarity, Local Frames Bag}

\maketitle

\section{Introduction}\label{sec1}

\begin{figure*}[t]
\centering
\includegraphics[trim=60 70 60 0, clip, width=0.95\linewidth]{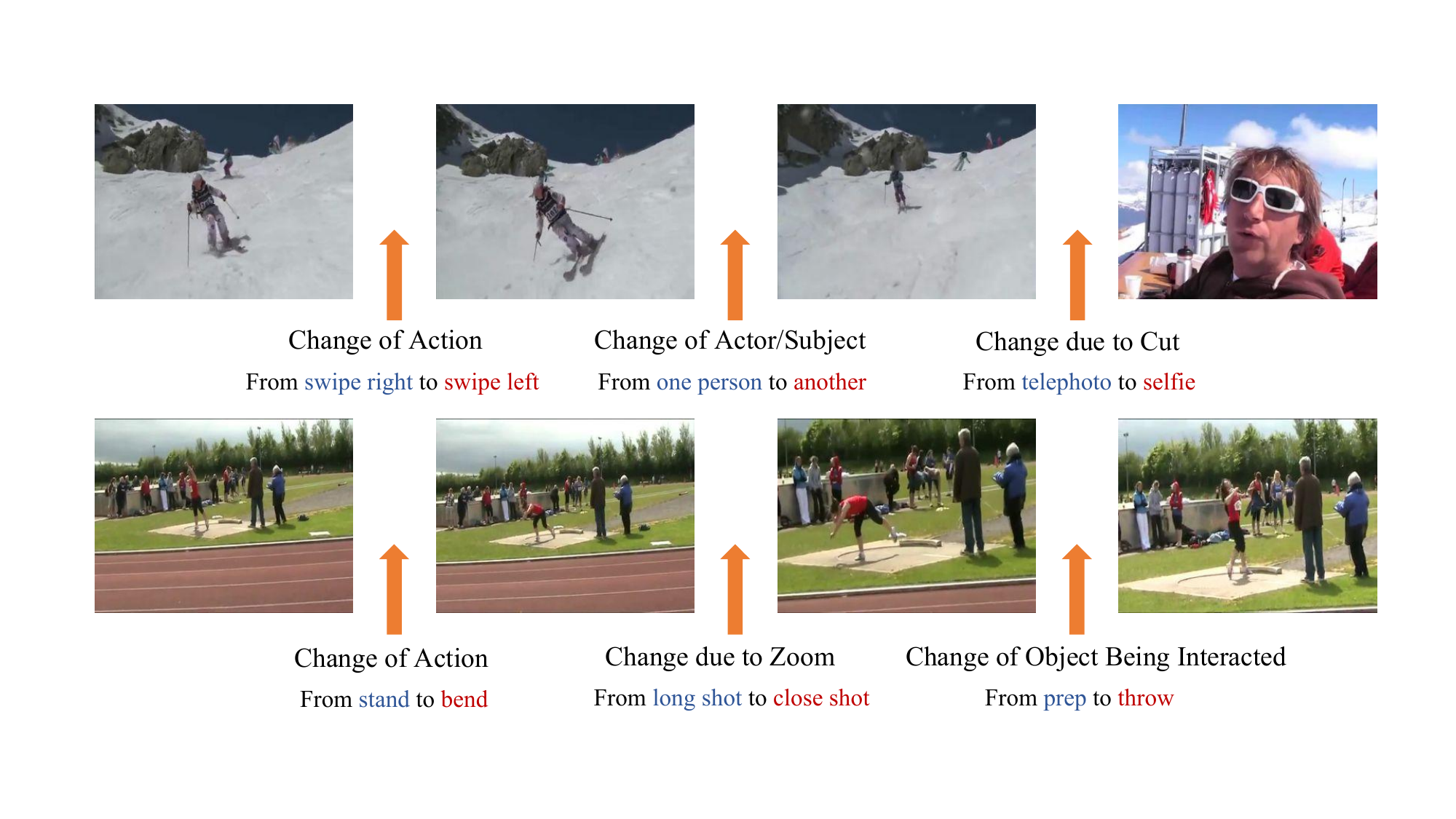}
\caption{Examples of generic event boundaries. The first video is segmented at a change of action, then the change of subject into another one and the change of a different scene. The second video is segmented due to action change, camera zoom and object interaction. The taxonomy-free nature of Kinetics-GEBD event boundaries makes it harder than existing tasks.}
\label{fig:gebd_example}
\end{figure*}

In recent years, video has become an integral part of human life, significantly impacting various aspects of our daily routines and activities. By 2023, the video content will make up 80\% of all consumer internet traffic\footnote{https://www.meltycone.com/blog/video-marketing-statistics-for-2023 ~\\~~~~$\dagger$ Equal contribution.}. When perceiving video contents, people will naturally and spontaneously segment events, breaking down longer events into a series of shorter temporal units \cite{DBLP:journals/corr/GEBD}. However, this mechanism is tough for machine learning, although it is so natural to the human brain. To this end, Generic Event Boundary Detection \cite{DBLP:journals/corr/GEBD} (GEBD) is proposed to allow machines to develop such an ability.

GEBD aims to localize the moments in which humans naturally perceive event boundaries. The high-level causes of event boundaries in the GEBD task are the following: \textit{ 1) change in spatial domain}: significant changes in the color or brightness of the environment. \textit{2) change in the temporal domain}: an old action ends or a new action starts. Notably, theses causes can happen simultaneously or are intermingled together, which lead to complicated event boundary variations, as shown in Figure~\ref{fig:gebd_example}. To solve the GEBD task, we can simply regard it as a video representation learning problem following the main methods. Currently, the two-stream networks \cite{DBLP:conf/nips/Two-Stream-Conv,DBLP:conf/cvpr/FeichtenhoferPZ16,DBLP:conf/iccv/Feichtenhofer0M19} and 3D convolutional networks \cite{DBLP:conf/eccv/TaylorFLB10,DBLP:journals/pami/JiXYY13,DBLP:conf/iccv/TranBFTP15,DBLP:journals/pami/VarolLS18} are two popular network architectures in the video understanding field. The two-stream networks usually incorporate two different modalities of information to learn complementary representation, for example, decoded RGB video frames and optical flow. 3D convolutional network is another choice to model temporal information using the spatio-temporal filters. Transformers were successfully applied in computer vision \cite{DBLP:conf/iclr/ViT},  the new trend in video understanding is using the Transformers, including \cite{DBLP:conf/iclr/DosovitskiyB0WZ21,DBLP:journals/corr/abs-2103-15691,DBLP:journals/corr/abs-2106-13230,DBLP:journals/corr/abs-2104-11227,DBLP:conf/mm/ZhangHN21}, which achieve competitive results. Despite their success, these methods are not optimal for the GEBD task since consecutive decoded RGB frames contain high temporal redundancy and are not practical for real-time applications.

Recently, another alternative for video understanding is learning directly from compressed domain. Several methods \cite{DBLP:conf/cvpr/ZhangWW0W16,DBLP:conf/mm/SIFP,DBLP:conf/cvpr/coviar,DBLP:conf/cvpr/dmc-net,DBLP:conf/iccv/WangGLD19,DBLP:conf/iclr/YuLKS21,DBLP:conf/cvpr/HuangLWPXJ21} have demonstrated the advantages of directly taking compressed information in video stream as input for video understanding. These methods usually run in two orders of magnitude faster than the methods using optical flow while achieving competitive results \cite{DBLP:conf/cvpr/dmc-net}. This tremendous improvement in speed comes from the using of motion vectors and residuals, which designed for storage and transmission of videos and almost compute-free. The rich information in motion vectors can be regarded as an alternative to the compute-intensive optical flow. To better utilize motion vectors and residuals, different methods have been developed for efficient and effect compressed video representation learning. Specifically, CoViAR \cite{DBLP:conf/cvpr/coviar} first converts motion vectors and residuals into 2D representations like images and then directly feeds them into 2D CNNs for action recognition. This method lacks interactions between I-frames and P-frames and thus achieves inferior results. DMC-Net~\cite{DBLP:conf/cvpr/dmc-net} improves the CoViAR~\cite{DBLP:conf/cvpr/coviar} method by reconstructing the optical flow based on motion vectors and residuals and a discriminator is applied to guide the reconstruction. However, it still needs optical flow in the training stage. SIFP \cite{DBLP:conf/mm/SIFP} uses the slow I pathway receiving a sparse sampling I-frame clip and the fast P pathway receiving a dense sampling pseudo-optical flow clip, which eliminates the dependence on traditional optical flows calculated from raw videos. Although the aforementioned method achieves promising results, they are still far from satisfactory, which lack effective fusion strategies between different modalities, such as decoded I-frames, motion vectors, and residuals. GEBD task is more sensitive to the local temporal context, which needs a new mechanism to learn from compressed information. 

\begin{figure}[t]
  \centering
   \includegraphics[width=1.0\linewidth]{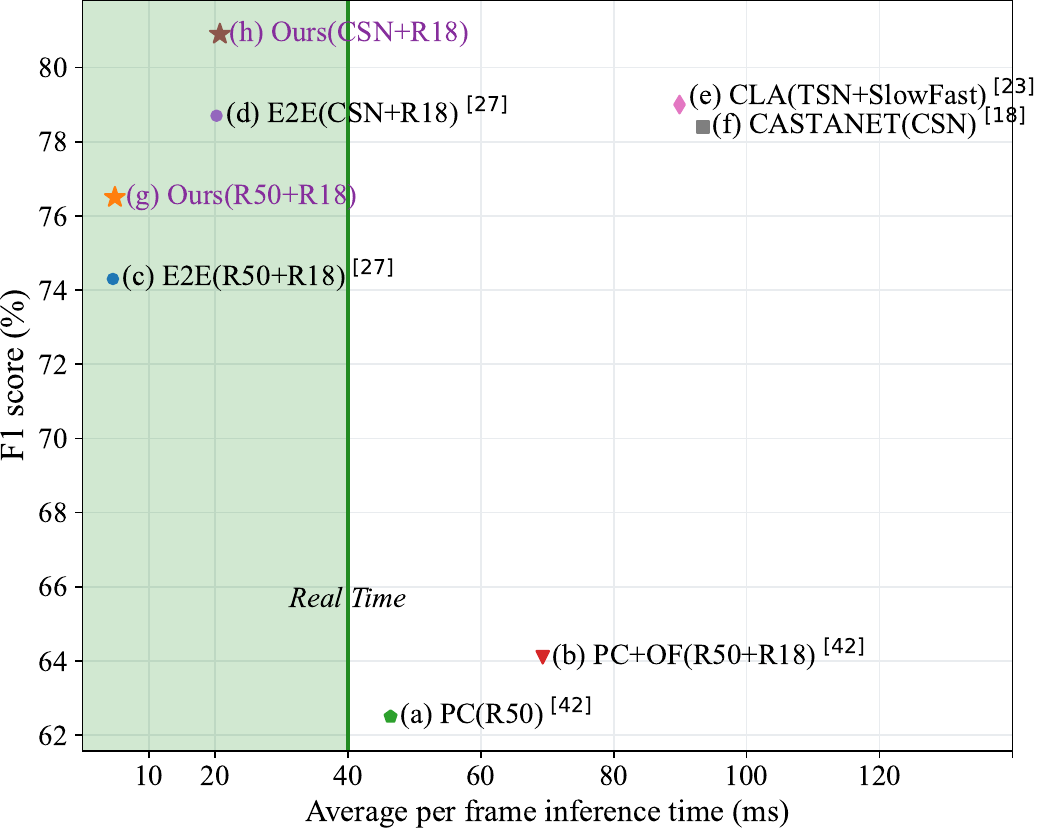}
   \caption{The inference time \textit{vs.} F1 score of different methods on the Kinetics-GEBD validation dataset \cite{DBLP:journals/corr/GEBD}. \textbf{(a)} The previous method \cite{DBLP:journals/corr/GEBD} PC is relatively fast with inferior results. \textbf{(b)} After integrating the optical flow (OF) module, the accuracy is improved with much slower running speed. \textbf{(c, d)} \cite{CVRL-GEBD-Li} leverages motion vectors and residuals in the compressed domain and achieves a competitive F1 score with a faster running speed. \textbf{(e, f)} CLA \cite{DBLP:journals/corr/abs-2106-11549} and CASTANET \cite{DBLP:journals/corr/abs-2107-00239} take fully decoded RGB frames as input, which are much slower than the methods conducted in compressed domain. \textbf{(g, h)} Compared with \textbf{(c, d)}, our method obtains about 2.0\% absolute improvements while keeping almost the same running speed. The green region indicates that the methods are run in real time. Best viewed in color for all figures throughout the paper.}
   \label{fig:onecol}
\end{figure}

In this paper, we focus on GEBD and develop a fast end-to-end method that can effectively learn from a local compressed video stream. The previous attempt \cite{DBLP:journals/corr/GEBD} formulate it as a classification task by considering the context information of the candidate boundaries. However, it neglects the temporal relations between consecutive frames and operates inefficiently during feature extraction stage. Inspired by \cite{DBLP:conf/cvpr/ZhangWW0W16,DBLP:conf/cvpr/coviar,DBLP:conf/cvpr/dmc-net,DBLP:conf/iccv/WangGLD19,DBLP:conf/iclr/YuLKS21,DBLP:conf/cvpr/HuangLWPXJ21}, we designed an end-to-end trained network to exploit discriminative features for GEBD in the compressed domain, \ie, MPEG-4, which can save decoding cost and improve feature extraction efficiency. Specifically, most modern codecs split a video into several group of pictures (GOP), where each GOP is formed by one I-frames and $T$ P-frames. To solve the difficulty that arose from the long chain of dependency of the P-frames, following CoViar~\cite{DBLP:conf/cvpr/coviar}, we use the backtracing technique to compute the accumulated motion vectors and residuals in linear time. In this way, the consecutive P-frames in each GOP depend only on the reference I-frame, which can be processed in parallel.

In contrast to the I-frame, it is difficult to learn the discriminative features of the P-frames. Refining the features of the reference I-frame based on the motion vectors and residuals becomes an intuitive option. Motion vectors and residuals provide information to reconstruct P-frames by referring the dependent I-frames. In addition to that, they also provide motion information that obtained from the video encoding process. To that end, we design a lightweight spatial-channel attention module to refine the features of the reference I-frame with the guidance of the motion vectors and residuals. In this way, the features of P-frames and I-frames are converted to the same feature space, which benefits the subsequent processing. After obtaining P-frame features, we split frame sequences into successive local frames bags. Each local frames bag only contains a fixed number of frames and is responsible for providing necessary context information to determine whether the central frame belongs to an event or not. Then we compute frame differences using group similarity in each local frames bag based on temporal context extracted by the long short-term memory (LSTM) module. This module can predict the event boundaries of videos accurately and it actually imitates humans, \ie, look back and forth around the candidate frames to determine event boundaries, by comparing the extracted features before and after the candidate frames. In addition, to remedy the ambiguities of annotations and speed up the training process, we use the Gaussian kernel to preprocess the ground-truth event boundaries instead of using the ``hard lables'' of boundaries. Extensive experiments conducted on the Kinetics-GEBD and TAPOS datasets to demonstrate the effectiveness of the proposed method. Specifically, the proposed method achieves comparable results to the state-of-the-art method at the CVPR'21 LOVEU Challenge \cite{DBLP:journals/corr/abs-2106-11549} with much faster running speed, as in Figure \ref{fig:onecol}.

In summary, we make the following contributions:

\begin{itemize}
    \item We propose a spatial-channel attention module (SCAM), which also refines P-frame features with I-frame features and shows advantages with bidirectional information flow compared to spatial-channel compressed encoder (SCCE).
    \item Instead of using linear temporal contrastive module, we split frame sequences into successive local frames bags and compute frame differences using group similarity in each local frames bag based on temporal context extracted by LSTM module.
    \item We provide additional ablation studies and qualitative analyses to demonstrate the superiority of each components comprehensively, including different backbones, different temporal feature extractor and learning scheme of different number of annotators.
    \item We achieve comparable results to the state-of-the-art methods at the CVPR'21 LOVEU Challenge \cite{DBLP:journals/corr/abs-2106-11549} with much faster running speed and obtains about 2\% absolute improvements compared with our preliminary version \cite{CVRL-GEBD-Li}, demonstrating its effectiveness. We also achieve 1.4\% improvement on TAPOS  \cite{DBLP:conf/cvpr/TAPOS} dataset compared to DDM-Net~\cite{DBLP:journals/corr/progressive}, which is developed on fully decoded RGB frames.
\end{itemize}

This paper is an extended version of a preliminary conference publication \cite{CVRL-GEBD-Li}. The main new contributions or differences include:

\begin{itemize}
    \item[1)] We improve the original SCCE by proposing the new SCAM that refines P-frame feature with I-frame features using bidirectional information flow.

    \item[2)] We propose to utilize the LSTM module to capture temporal information for better performance.

    \item[3)] We carry out more ablative studies to analyze each component of our approach in-depth.

    \item[4)] Notable performance gains are achieved with the aforementioned new contributions in comparison with our preliminary version in \cite{CVRL-GEBD-Li}.
\end{itemize}

The remainder of this paper is organized as follows. A brief review of related works is presented in Section II. The details of the proposed end-to-end method for generic event boundary detection and experimental explanations are given in Section III. Extensive experiments and ablation studies are given in Section IV. We conclude our method in Section V.

\section{Related Work}
\label{sec:related_work}

\subsection{Temporal Action Localization (TAL)}
TAL aims to localize the action segments from untrimmed videos. More specifically, for each action segment, the goal is to detect the start point, the end point and the action class it belongs to. Most approaches could be categorised into two groups, including two-stage methods \cite{richard2016temporal,ni2016progressively,caba2017scc,zhao2017temporal,Chao_2018_CVPR} and single-stage methods \cite{lea2017temporal,lin2017single,alwassel2018action,long2019gaussian,Yuan_2017_CVPR,ma2016learning,Yuan_2017_CVPR,zhao2020bottom}. In the two-stage method setting, the first stage generates action segment proposals. The actionness and the type of action for each proposal are then determined by the second stage, along with some post-processing methods such as grouping \cite{zhao2017temporal} and Non-maximum Suppression (NMS) \cite{DBLP:conf/iccv/BMN} to eliminate redundant proposals. For one-stage methods, the classification is performed on the pre-defined anchors \cite{lin2017single,long2019gaussian} or video frames \cite{ma2016learning,Yuan_2017_CVPR}.

TAL and GEBD are both tasks in video understanding, which solve the problem of boundary localization by extracting key information from videos. However, TAL is designed to locate specific action behaviors only, whereas GEBD can locate general events beyond action behaviors. In other words, TAL is a subtask within GEBD that can only locate actions. Research in TAL has provided many methods \cite{richard2016temporal,ni2016progressively,caba2017scc,lea2017temporal,lin2017single,alwassel2018action,long2019gaussian,Yuan_2017_CVPR,ma2016learning} for action localization, and we have addressed the GEBD task by extending these methods to generic event localization.

\subsection{Generic Event Boundary Detection}
The goal of GEBD \cite{DBLP:journals/corr/GEBD} is to localize the taxonomy-free event boundaries that break a long event into several short temporal segments. Different from Temporal Action Localization (TAL), GEBD only requires to predict the boundaries of each continuous segments. The current methods\cite{DBLP:journals/corr/abs-2106-11549,DBLP:journals/corr/abs-2107-00239,DBLP:journals/corr/abs-2106-10090} all follow the similar fashion in \cite{DBLP:journals/corr/GEBD}, which takes a fixed length of video frames before and after the candidate frame as input, and separately determines whether each candidate frame is the event boundary or not. Kang \etal \cite{DBLP:journals/corr/abs-2106-11549} use the temporal self-similarity matrix (TSM) as the intermediate representation and exploit the discriminative features with the popular contrastive learning approach for better performance. Hong \etal \cite{DBLP:journals/corr/abs-2107-00239} use the cascade classification heads and dynamic sampling strategy to boost both recall and precision. Rai \etal \cite{DBLP:journals/corr/abs-2106-10090} attempt to learn the spatiotemporal features using a two stream inflated 3D convolutions architecture. DDM-Net \cite{DBLP:journals/corr/progressive} present dense difference maps (DDM) to comprehensively characterize the motion pattern and exploit progressive attention on multi-level DDM to jointly aggregate appearance and motion clues. All these methods are developed on decoded RGB images and cannot benefit from almost compute-free motion vectors and residuals.

\subsection{Attention Mechanism}
Attention mechanism has been widely adopted in deep learning model design. The core of attention mechanism is to recalibrate the origin input feature with different weights in different dimensions. Transformer \cite{DBLP:conf/nips/att_is_all_you_need} and Non-local network \cite{DBLP:conf/cvpr/non-local} can capture long-range dependencies by computing the response at a spatial position as a weighted sum of the features at all positions in the input feature maps. SENet \cite{DBLP:conf/cvpr/SENet} develops the ``Squeeze-and-Excitation'' (SE) block that adaptively recalibrates channel-wise feature responses by explicitly modelling interdependencies between channels. In addition to channel-wise recalibration, CBAM \cite{DBLP:conf/eccv/CBAM} sequentially infers attention maps along both channel and spatial dimensions, and then uses the attention maps to recalibrate the origin input feature. In contrast to the aforementioned methods, we attempt to refine the P-frame feature with the guidance of motion vectors and residuals by considering both spatial and channel dimensions of the features of I-frame. Different from SCCE \cite{CVRL-GEBD-Li}, we also refine the origin P-frame feature with the guidance of reference I-frame feature. This bidirectional information flow can filter noises and fully leverages the information in compressed video stream.

\begin{figure*}[t]
\centering
\includegraphics[width=\textwidth]{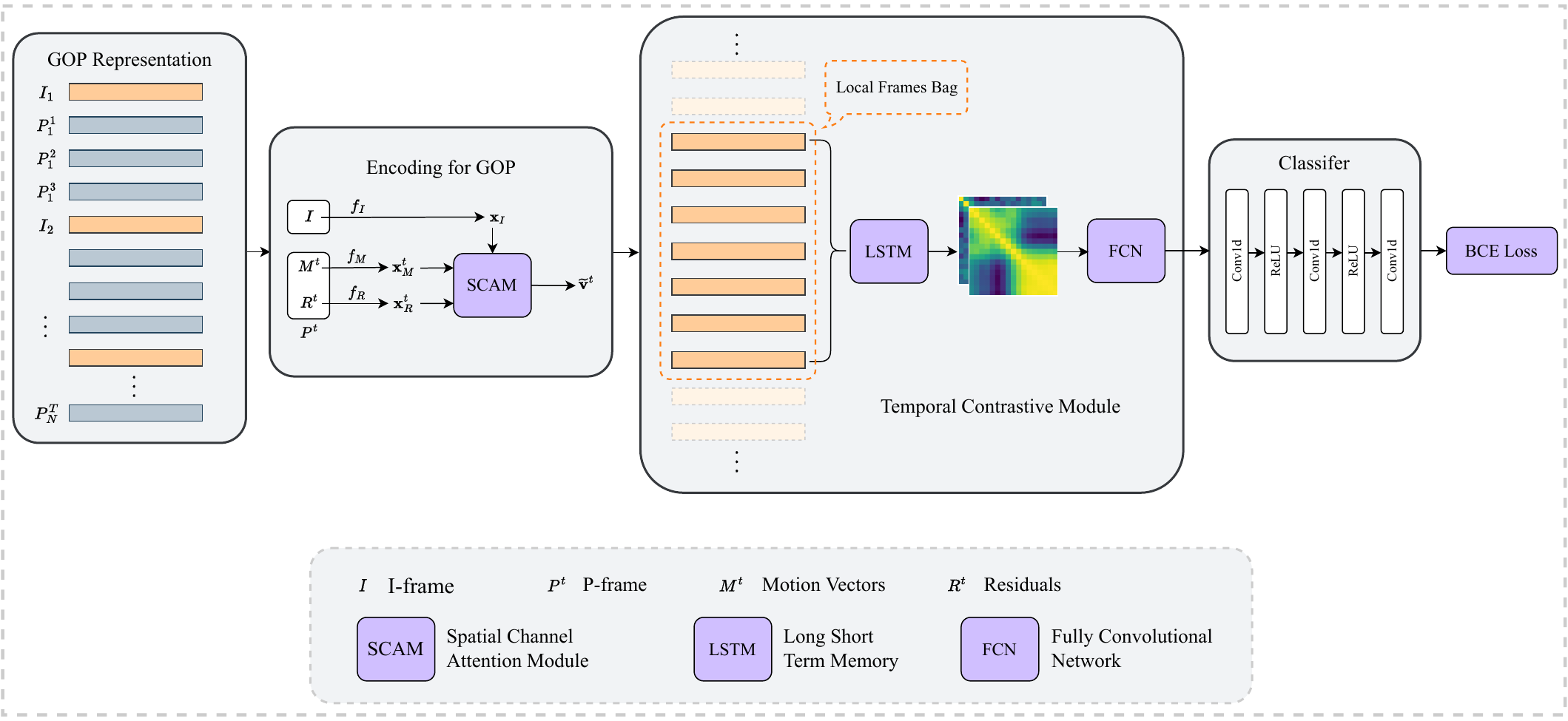}
\caption{The architecture of our method. The spatial-channel attention module (SCAM) is designed to obtain the refined P-frame representation $\widetilde{\bf v}^t$ based on reference I-frame feature ${\bf x}_I$, resized motion vectors $M^t$ and resized residuals $R^t$. This module regards each GOP as a process unit, which is efficient and can be paralleled in a large batch size. Then we use temporal contrastive module to capture temporal dependence explicitly based on unified representation $\widetilde{\bf v}^t$, which provides strong cues for boundary detection. After that, a simple classifier is applied to make the final predictions trained with the Gaussian smoothed soft labels.}
\label{fig:overall_architecture}
\end{figure*}

\section{Method}
\label{sec:method}
The existing method \cite{DBLP:journals/corr/GEBD} formulates the GEBD task as binary classification, which predicts the boundary labels for each frame by considering the temporal context information. That is, The preceding and succeeding frames of each video frame are feed into a neural network to detect the boundaries. It is inefficient due to the duplicated computation conducted in consecutive frames. To remedy this, we propose an end-to-end compressed video representation method for GEBD, which regards each video clip as a whole. Specifically, we use MPEG-4 encoded videos as our input. Each video clip ${\cal V}$ is formed by $N$ groups of pictures (GOPs), and each GOP contains one I-frame and $T$ P-frames, \ie,
\begin{equation}
\begin{array}{ll}
    {\cal V}=\big\{I_i, P_i^1, P_i^2, \cdots, P_i^T\big\}_{i=1}^N,
\end{array}
    \label{equ:video_representation}
\end{equation}
where $I_i \in \mathbb{R}^{3 \times {\cal H} \times {\cal W}}$ denotes the reference I-frame and $P_i^t$ denotes the $t$-th P-frame of the $i$-th GOP, and ${\cal H}$ and ${\cal W}$ are the height and width of the video frame. For simplicity, we assume that there exists the same number of P-frames in all GOPs. The assumption of a fixed number of P-frames simplifies the process, as it allows us to standardize the temporal length of each video segment for processing.
The P-frame $P_i^t$ in the $i$-th GOP is formed by the initial motion vector ${\cal M}_i^t \in \mathbb{R}^{2 \times {\cal H} \times {\cal W}}$ and initial residual ${\cal R}_i^t \in \mathbb{R}^{3 \times {\cal H} \times {\cal W}}$, which can be firstly obtained nearly cost-free from the compressed video stream, and then trace all motion vectors back until to the reference I-frame and accumulate the residual on the way to decouple the dependencies between the consecutive P-frames. In this way, each P-frame only depends on the reference I-frame rather than other P-frames. After that, we build our model based on the backtraced motion vectors and residuals and regard each GOP as a process unit. The overall network architecture is presented in Figure \ref{fig:overall_architecture}. As shown in Figure \ref{fig:overall_architecture}, the GOP is first encoded by the designed spatial-channel attention module (SCAM) to generate the unified video representation. After that, a temporal contrastive module is used to exploit the temporal context information to obtain the discriminative feature representations. Finally, a classifier is used to generate the accurate event boundaries. Our algorithm flow is shown in Alg. \ref{algo:cmd}.

\begin{algorithm}[t]
\centering
\caption{Local Compressed Video Stream Learning for GEBD}\label{algo:cmd}
\noindent\textbf{Require:} Input the I-frame $I$, $T$ motion vectors ${\cal M}^t$ and residuals ${\cal R}^t$, $l$ is the position of the candidate frame in the time sequence, and $k$ is the number of adjacent frames.
\begin{algorithmic}[1]
\State ${\bf x}_{I} = \operatorname{ResNet-50}(I)$ 
\For{$t = 1, \cdots, T$}
    \State ${\bf x}_{M}^t = \operatorname{ResNet-18}({\cal M}^t)$
    \State ${\bf x}_{R}^t=\operatorname{ResNet-18}({\cal R}^t)$
    \State $\widetilde{\bf v}^t = SCAM({\bf x}_{I}, {\bf x}_{M}^t, {\bf x}_{R}^t)$
\EndFor
\State $\mathcal{B}^l = \{\widetilde{\bf v}^{l-k}, \cdots, \widetilde{\bf v}^{l}, \cdots, \widetilde{\bf v}^{l+k}\}$
\State $\widetilde{\mathcal{B}}^l = LSTM(\mathcal{B}^l)$
\State $\mathcal{S}^l = \operatorname{Group-Similarity}(\widetilde{\mathcal{B}}^l, \widetilde{\mathcal{B}}^l)$
\State $\mathcal{P}^l = \operatorname{Average-Pool}(\operatorname{FCN}(\mathcal{S}^l))$
\State $\mathcal{Y}^l = \operatorname{Classifer-Head}(\mathcal{P}^l)$
\State ${\bf return}~\mathcal{Y}^l$
\end{algorithmic}
\end{algorithm}

\subsection{Learning from Spatial-Channel Attention Module}

\begin{figure}[t]
  \centering
   \includegraphics[width=1.0\linewidth]{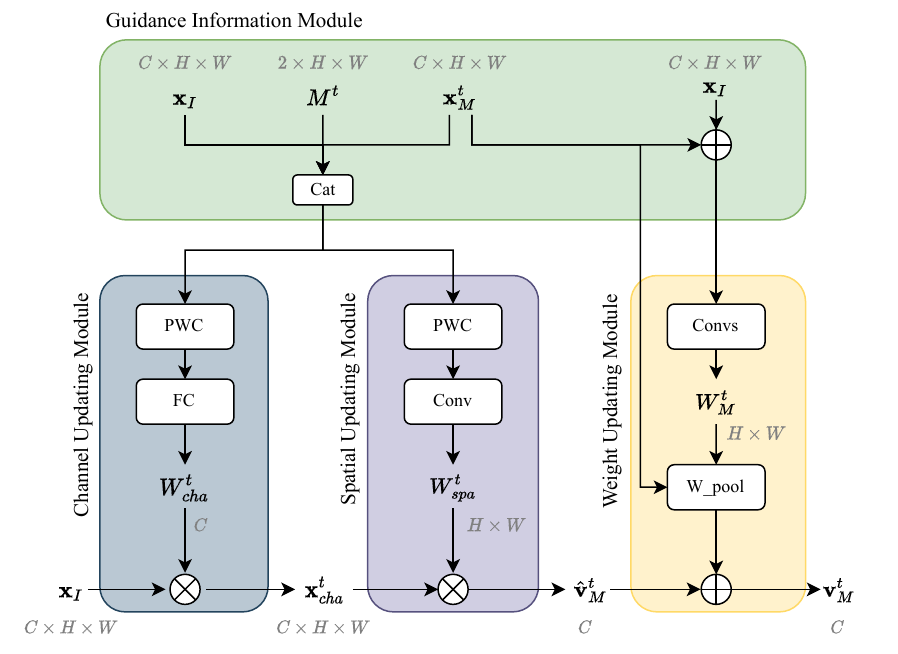}
   \caption{The architecture of the proposed spatial-channel attention module (SCAM). W\_pool corresponds to the weighted pool function. We concatenate the features of I-frame ${\bf x}_I$, resized motion vectors $M^t$ and the features of motion vectors ${\bf x}_M^t$ to modulate the features of the reference I-frame ${\bf x}_I$ in both channel and spatial dimensions. After that, we also use I-frame ${\bf x}_I$ feature to modulate the origin P-frame feature ${\bf x}_M^t$ with spatial weight $W^t_M$. Different from the previous method \cite{CVRL-GEBD-Li} where the modulated features $\hat{{\bf v}}_M^t$ is residually added with the features of motion vectors, our method can learn more robust representation with the bidirectional information flow.}
   \label{fig:scce}
\end{figure}

\label{sec:deformable_updating}
Motion, uncovered regions, and lighting variations frequently happen in video sequences. Modern codecs use macroblock as the basic unit for motion-compensated prediction in a number of mainstream visual coding standards such as MPEG-4, H.263, and H.264. Motion vectors record the moving direction of each macroblock with respect to its reference frame(s), describing the motion patterns of videos, which is important for the GEBD task. The residuals can be regraded as the compensations of the motion information, which contains the boundary information of moving objects and plays a crucial role in identifying the important regions in the I-frame. Thus, we propose applying the attention mechanism to different regions of I-frame with the guidance of motion vectors to enrich the features by considering both channel and spatial dimensions. For simplicity, we omit the index $i$ of the GOP in the following sections.

Firstly, we use the convolutional neural network taking the decoded RGB image as input to extract the feature representation ${\bf x}_{I}$ of the I-frame $I$, \ie, ${\bf x}_{I} = f_I(I)$, where ${\bf x}_I \in \mathbb{R}^{C \times H \times W}$ is the features of the I-frame $I$, and $C$, $H$ and $W$ are the channel, height and with of the features ${\bf x}_I$, respectively. $f_I(\cdot)$ denotes the model used to extract features for the I-frame, which is pre-trained on large-scale datasets (\eg, ResNet50 pre-trained on ImageNet). Meanwhile, we can similarly compute the features for the P-frames $\{P^1, P^2, \cdots, P^T\}$ with ResNet-18, by directly taking the initial motion vectors ${\cal M}^t$ and initial residuals ${\cal R}^t$, \ie, ${\bf x}_{M}^t = f_M({\cal M}^t)$, and ${\bf x}_{R}^t=f_R({\cal R}^t)$ as input, where ${\bf x}_{M}^t, {\bf x}_{R}^t \in \mathbb{R}^{C \times H \times W}$ denote the features of the motion vectors and residuals, respectively. 

We choose the ResNet-18 to extract motion vectors and residuals features for two reasons: (1) Motion vectors and residuals, as types of compressed-domain information, are typically characterized by an uneven distribution with most positions being empty and only a few containing meaningful data. Therefore, for such sparse information, using a small network (such as ResNet-18) to extract features may be more effective in processing it. (2) ResNet-18 has a powerful generalization capability and the success it has achieved in many image analysis tasks. Although motion vectors and residuals may not directly correlate with natural images, the key concept is to leverage the high-level feature extraction ability of the pre-trained network to create useful representations. 
In this way, a considerable amount of time can be saved on extracting features for the P-frames. This simple strategy \cite{DBLP:conf/cvpr/coviar}, in which the motion vector and residual are treated as separate branches without fusion, can only provide limited performance improvement.
The method \cite{DBLP:conf/cvpr/dmc-net} attempts to integrate the optical flow in the training phase, which can further improve the accuracy. However, there is still much room for improvement of the aforementioned methods. Specifically, the motion vectors record the motion patterns of both the scenes and objects in videos, and the residuals provide the compensation information. Both of them do not contain the context information of the scenes. To this end, we design the spatial channel compressed encoder module by integrating the features of the reference I-frame $x_I$ in computing the features of P-frames.  

We first compute the features ${\bf x}_{M}^t$ of the motion vectors by refining the features of the reference I-frame ${\bf x}_I$ in both the channel and spatial dimensions. As indicated by \cite{DBLP:conf/eccv/ZeilerF14}, different regions on the feature maps focus on different parts of the images. Thus, we introduce the attention weight for each feature map of ${\bf x}_I$ based on the information of the P-frame ${\bf x}_{M}^t$. Specifically, we concatenate I-frame feature ${\bf x}_I$, motion vector feature ${\bf x}_{M}^t$ and resized motion vectors $M^t$ (resized from ${\cal M}^t$) together in the channel dimension to compute the channel weight $W_\text{cha}^t$ using a lightweight PWC-Net \cite{DBLP:conf/cvpr/SunY0K18}. PWC-Net is chosen to maintain consistency with previous work \cite{DBLP:conf/cvpr/dmc-net}, and because of the similarity between motion vectors and optical flow information, which makes it suitable for processing using an optical flow network. \ie,

\begin{equation}
\begin{array}{ll}
    {\bf z}_\text{cha}^t &= \operatorname{PWC}([{\bf x}_I; {\bf x}_{M}^t; M^t]) \\
    {\bf h}_\text{cha}^t &= \operatorname{avg\_pool}({\bf z}_\text{cha}^t) \\
    W_\text{cha}^t &= \sigma(W_2 \cdot \zeta(W_1{\bf h}_\text{cha}^t+b_1)+b_2)
\end{array}
\end{equation}
where $\sigma$ is the sigmoid function, $\zeta$ is the ReLU function, and $W_1, b_1, W_2, b_2$ are the learnable weights of the FC layers. After that, the features of the I-frame ${\bf x}_I$ are updated based on $W_\text{cha}^t$ as follows,
\begin{equation}
\begin{array}{ll}
    {\bf x}_\text{cha}^t = {\bf x}_I \otimes W_\text{cha}^t
\end{array}
\end{equation}
where $\otimes$ is the channel-wise multiplication. In this way, we can compute the channel-weighted feature ${\bf x}_\text{cha}^t$ by updating ${\bf x}_I$ in channel dimension, with the guidance of the motion vectors. Meanwhile, the channel-weighted feature ${\bf x}_\text{cha}^t$ is further updated in the spatial dimension and the spatial dimension is reduced. That is, given the features ${\bf x}_I$ of the reference I-frame, motion vector features ${\bf x}_{M}^t$ and resized motion vectors $M^t$, we compute the 2D weight map $W_\text{spa}^t$, \ie, 

\begin{equation}
\begin{array}{ll}
    {\bf z}_\text{spa}^t &= \operatorname{PWC}([{\bf x}_I; {\bf x}_{M}^t; M^t]) \\
    {\bf h}_\text{spa}^t &= \operatorname{2d\_conv}({\bf z}_\text{spa}^t) \\
    W_\text{spa}^t &= \operatorname{softmax}({\bf h}_\text{spa}^t)
\end{array}
\end{equation}
where $W_\text{spa}^t \in \mathbb{R}^{H \times W}$ is the spatial weight map. The softmax function is applied to $h^t_{spa}$ across a 2D spatial map, guaranteeing that the sum of values at each spatial location in the feature map equals 1. After that, we use $W_\text{spa}^t$ to weight the features ${\bf x}_\text{cha}^t$ in the spatial dimension to compute the enriched features of the motion vectors $\hat{{\bf v}}_M^t \in \mathbb{R}^C$, \ie, 

\begin{equation}
\begin{array}{ll}
    \hat{{\bf v}}_M^t = \sum_{i=1}^H\sum_{j=1}^W{{\bf x}_\text{cha}^t(i,j) \cdot W_\text{spa}^t(i,j)}
\end{array}
\end{equation}
where $i,j$ are the spatial positions of ${\bf x}_\text{cha}^t$ and $W_\text{spa}^t$. Previous method \cite{CVRL-GEBD-Li} obtains the refined features of the motion vectors ${\bf v}_M^t \in \mathbb{R}^C$ use residual addition, which doesn't consider the information flow from I-frame feature to P-frame feature. Thus we propose to modulate P-frame feature in spatial dimension by using weighted pooling method. The method applies the weight value $W^t_M$ (shape of $H\times W$) to each channel at the same position in $x^t_M$ (shape of $C\times H\times W$), thereby performing the weighted pooling, \ie,

\begin{equation}
    {\bf v}_M^t = \hat{{\bf v}}_M^t + \operatorname{weighted\_pool}(\operatorname{Convs}({\bf x}_M^t + {\bf x}_I))
\end{equation}
where $\operatorname{Convs}$ is a 3-layer Conv-ReLU network and the output channel of last layer is 1 for predicting a 2D spatial weight map. The overall computing process of ${\bf v}_M^t$ is presented in Figure \ref{fig:scce}. Similarly, we can compute the refined features for the residuals ${\bf v}_R^t \in \mathbb{R}^C$. The final feature representations for the P-frame is further computed as
\begin{equation}
    \widetilde{\bf v}^t = {\bf v}_M^t + {\bf v}_R^t
\end{equation}

In this way, we can compute the features of the P-frames $\{\widetilde{\bf v}^1, \widetilde{\bf v}^2, \cdots, \widetilde{\bf v}^T\}$ in the GOP by considering the reference I-frame $I$ in both channel and spatial dimensions. The overall process is very efficient and can be processed in parallel in GOPs. After extracting the discriminative features for both the I-frames and P-frames in the same feature space, we can predict the event boundaries efficiently and accurately.

\subsection{Learning from Local Frames Bag}

\begin{figure*}[t]
\centering
\includegraphics[width=0.95\linewidth]{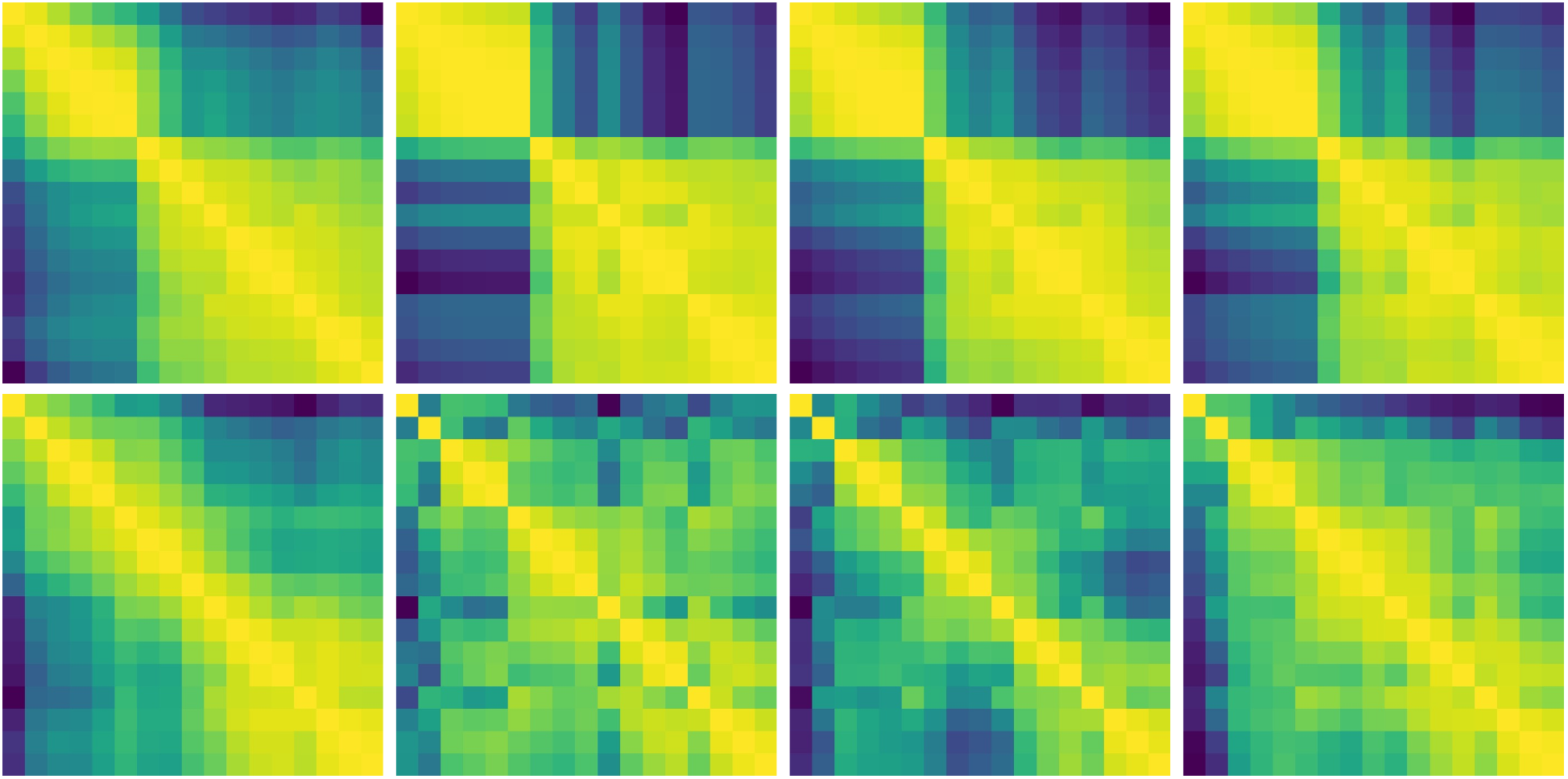} 
\caption{Visualization of grouped similarity maps $\mathbf{S}_t$, $G=4$ in this example. First row indicates that there is a potential boundary in this local sequence while the second row shows no boundary in this sequence. We can also observe slightly different patterns between the same group, which may imply that each group is learning in a different aspect.}
\label{fig:similarity_maps}
\end{figure*}

\label{sec:temporal_parser}
Based on the extracted features of the video ${\cal V}$, we aim to design a temporal module to predict the event boundaries accurately. The existence of an event boundary in a video clip implies that there is a visual content change at that point, thus it is very difficult to infer the boundary from one single frame. As a result, the key clue for event boundary detection is to localize changes in the temporal domain. Inspired by humans, \ie, look back and forth around the candidate boundary frames to determine event boundaries, we construct a local frames bag for each candidate frame and each local frames bag is responsible for providing context information to predict an event boundary. Specifically, for each candidate frame $\widetilde{\bf v}^{l}$ at time $l$, we construct a local frames bag for it by gathering adjacent $k$ frames before candidate frame $\widetilde{\bf v}^{l}$ and adjacent $k$ frames after $\widetilde{\bf v}^{l}$, resulting in a local frames sequence, namely local frames bag:
\begin{equation}
\begin{array}{l}
     \mathcal{B}^l = \{\widetilde{\bf v}^{l-k}, \widetilde{\bf v}^{l-(k-1)}, \cdots, \widetilde{\bf v}^{l}, \cdots, \widetilde{\bf v}^{l+(k-1)}, \widetilde{\bf v}^{l+k}\},
\end{array}
\end{equation}
which can be implemented through efficient memory view method provided by modern deep learning framework and processed in parallel.

After obtaining local frames bag $\mathcal{B}^l$ for candidate frame $\widetilde{\bf v}^{l}$, we use a 2-layer long short-term memory (LSTM) to learn temporal relationships. Formally, for each frame $\widetilde{\bf v}^{t}$ in local frames bag $\mathcal{B}^l$, we compute the hidden state in each layer as follows:
\begin{equation}
\begin{array}{ll}
            i^t = \sigma(W_{ii} \widetilde{\bf v}^{t} + b_{ii} + W_{hi} h_{t-1} + b_{hi}) \\
            f^t = \sigma(W_{if} \widetilde{\bf v}^{t} + b_{if} + W_{hf} h_{t-1} + b_{hf}) \\
            g^t = \tanh(W_{ig} \widetilde{\bf v}^{t} + b_{ig} + W_{hg} h_{t-1} + b_{hg}) \\
            o^t = \sigma(W_{io} \widetilde{\bf v}^{t} + b_{io} + W_{ho} h_{t-1} + b_{ho}) \\
            c^t = f^t \odot c^{t-1} + i^t \odot g^t \\
            h^t = o^t \odot \tanh(c^t), \\
\end{array}
\end{equation}
where $h^t$ is the hidden state at time t, $c^t$ is the cell state at time t, $\widetilde{\bf v}^{t}$ is the input frame at time t, $h^{t-1}$ is the hidden state of the layer at time $t-1$ or the initial hidden state at time $0$, and $i^t, f^t, g^t, o^t$ are the input, forget, cell, and output gates, respectively. $\sigma$ is the sigmoid function, and $\odot$ is the Hadamard product. While there are more options to learn temporal relations like famous Transformer~\cite{DBLP:conf/nips/attisallyouneed}, we use a simple 2-layer LSTM which works very well as shown in Table~\ref{tab:ablation_temporal}.

After learning the temporal relationships, we obtain temporal-enhanced local frames bag $\widetilde{\mathcal{B}}^l = \{h^{l-k}, h^{l-(k-1)}, \cdots, h^{l}, \cdots, h^{l+(k-1)}, h^{l+k}\}$. The LSTM module aims at discovering relationships between frames and giving high level representation of frames sequences. However, event boundaries emphasize the differences between adjacent frames and neural networks tend to take shortcuts during learning \cite{DBLP:journals/natmi/GeirhosJMZBBW20}. Thus classifying these frames directly into boundaries may lead to inferior performance due to non-explicit cues. Based on this intuition, we propose to guide classification with feature similarity of each frame pair in the local frames bag $\widetilde{\mathcal{B}}^l$. Instead of performing similarity calculation with all $C$-dimensional channels, we found it beneficial to split the channels into several groups and calculate the similarity of each group independently. Formally, the group similarity map $\mathcal{S}^l$ is calculated as follows. First, split $\widetilde{\mathcal{B}}^l$ in the channel dimension to get $\{\widetilde{b}^l_i\}_{i=1}^{i=G}$, where $\widetilde{b}^l_i \in \mathbb{R}^{{C \over G} \times (2k+1) \times (2k+1)}$ and $G$ is the number of groups. Second, we calculate the cosine similarity of each $\widetilde{b}^l$ as follows:
\begin{equation}
    \begin{array}{l}
         \mathcal{S}^l = [\operatorname{sim}(\widetilde{b}^l_1, \widetilde{b}^l_1), ..., \operatorname{sim}(\widetilde{b}^l_G, \widetilde{b}^l_G)],
    \end{array}
\end{equation}
where $\mathcal{S}^l \in \mathbb{R} ^{G \times (2k+1) \times (2k+1)}$. As the group similarity map $\mathcal{S}^l$ contains the similarity scores, it shows different patterns (as shown in Figure \ref{fig:similarity_maps}) in different sequences, which are critical for boundary detection. Then we use a 4-layer fully convolutional network \cite{DBLP:conf/cvpr/FCN} to learn the similarity patterns, which we found works very well and efficient enough. Then we average pool the output of FCN to get a vector representation $\mathcal{P}^l$, and this vector is used for final event boundary classification:
\begin{equation}
    \begin{array}{l}
         \mathcal{P}^l = \operatorname{average-pool}(\operatorname{FCN}(\mathcal{S}^l)),
    \end{array}
\end{equation}
where $\mathcal{P}^l \in \mathbb{R}^{C}$. Then for the final classification, we use the contrastive representations $\{ \mathcal{P}^1,  \mathcal{P}^2, \cdots,  \mathcal{P}^T\}$ to make the event boundary predictions.

\subsection{Loss Function}
Given feature representations $\{ \mathcal{P}^1,  \mathcal{P}^2, \cdots,  \mathcal{P}^T\}$ of each video frame and the corresponding ground-truth labels, the event boundary detection task is intuitively formulated as the binary classification task. However, the ambiguities of annotations disrupt the learning process, which leads to poor convergence. To solve this issue, we use the Gaussion kernel to preprocess the ground-truth event boundaries to obtain the soft labels instead of using the ``hard labels'' of boundaries. Specifically, for each annotated boundary, the intermediate label of the neighboring position $i$ is computed as:
\begin{equation}
    g_i^l = \exp\Big( -\frac{( l-i )^2}{2\alpha^2} \Big)
\end{equation}
where $g_i^l$ indicates the intermediate label at time $i$ corresponding to the annotated boundaries at time $l$. We set $\alpha =1$ in all our experiments. The final soft labels are computed as the summation of all intermediate labels. Finally, a simple nonlinear Conv1D classifier is applied to predict the boundary score $S^l$ and the binary cross-entropy loss is used to guide the training process.

\section{Experiments}
\label{sec:experiments}

\renewcommand{\arraystretch}{1.05}
\begin{table*}[t]\small
\caption{The evaluation results on the Kinetics-GEBD validation set with different Rel.Dis. thresholds. }
\centering
\setlength{\tabcolsep}{1.45mm}
 \begin{tabular}{l|cccccccccc|c}
\toprule
Rel.Dis. Threshold & 0.05  & 0.1  & 0.15  & 0.2 & 0.25 & 0.3  & 0.35 & 0.4 & 0.45 & 0.5 &avg \\
\midrule
BMN \cite{DBLP:conf/iccv/LinLLDW19} & 0.186  & 0.204 & 0.213 & 0.220 & 0.226 & 0.230 & 0.233 & 0.237 & 0.239 & 0.241 &0.223 \\
BMN-StartEnd \cite{DBLP:journals/corr/GEBD} & 0.491 & 0.589 & 0.627 & 0.648 & 0.660 & 0.668 & 0.674 & 0.678 & 0.681 &0.683 &0.640 \\
TCN-TAPOS \cite{DBLP:journals/corr/GEBD} &0.464  &0.560  &0.602  &0.628  &0.645  &0.659  &0.669  &0.676  &0.682  &0.687 &0.627 \\
TCN \cite{DBLP:conf/eccv/LeaRVH16} & 0.588 & 0.657 & 0.679 & 0.691 & 0.698 & 0.703 & 0.706 & 0.708 & 0.710 &0.712 &0.685   \\
PC \cite{DBLP:journals/corr/GEBD}  & 0.625 &0.758 &0.804 &0.829 &0.844 &0.853 &0.859 &0.864 &0.867 &0.870 &0.817 \\
PC + Optical Flow &0.646 & 0.776 & 0.818 & 0.842 & 0.856 & 0.864 & 0.868 & 0.874 & 0.877 & 0.879 & 0.830 \\
E2E \cite{CVRL-GEBD-Li}  &0.743&0.830 & 0.857 & 0.872 & 0.880 &0.886  & 0.890 & 0.893 & 0.896 &0.898 & 0.865\\
DDM-Net~\cite{DBLP:journals/corr/progressive} &0.764&0.843 & 0.866 & 0.880 & 0.887 &0.892  & 0.895 & 0.898 & 0.900 &0.902 & 0.873\\
\midrule
Ours &\textbf{0.768}&\textbf{0.848} & \textbf{0.872} & \textbf{0.885} & \textbf{0.892} &\textbf{0.896}&\textbf{0.899}& \textbf{0.901} & \textbf{0.903} &\textbf{0.906} & \textbf{0.877}\\
\bottomrule
\end{tabular}
\label{tab:results_of_kinetics_val}
\end{table*}

\begin{table*}[t]\small
\centering
\setlength{\tabcolsep}{1.5mm}
\caption{The evaluation results on the TAPOS validation set with different Rel.Dis. thresholds.}
 \begin{tabular}{l|cccccccccc|c}
\toprule
Rel.Dis. threshold& 0.05 & 0.1 & 0.15 & 0.2 & 0.25 & 0.3 & 0.35 & 0.4 & 0.45 & 0.5 & avg \\ 
\midrule
ISBA~\cite{DBLP:conf/cvpr/ISBA} &0.106&0.170&0.227&0.265&0.298&0.326&0.348&0.369&0.382&0.396&0.302\\
TCN~\cite{DBLP:conf/eccv/LeaRVH16} & 0.237 & 0.312 & 0.331 & 0.339 & 0.342 & 0.344 & 0.347 & 0.348 & 0.348 & 0.348 & 0.330\\
CTM~\cite{DBLP:conf/eccv/CTM} & 0.244 & 0.312 & 0.336 & 0.351 & 0.361 & 0.369 & 0.374 & 0.381 & 0.383 & 0.385 & 0.350\\
TransParser~\cite{DBLP:conf/cvpr/TAPOS}& 0.289 & 0.381 & 0.435 & 0.475 & 0.500 & 0.514 & 0.527 & 0.534 & 0.540&0.545&0.474 \\
PC~\cite{DBLP:journals/corr/GEBD}& 0.522 & 0.595 & 0.628 & 0.646 & 0.659 & 0.665 & 0.671 & 0.676 & 0.679 & 0.683 & 0.642 \\
DDM-Net~\cite{DBLP:journals/corr/progressive}& 0.604 & 0.681 & 0.715 & 0.735 & 0.747 & 0.753 & 0.757 & 0.760 & 0.763 & 0.767 & 0.728 \\
\midrule
Ours &\textbf{0.618} &\textbf{0.694} &\textbf{0.728} &\textbf{0.749} &\textbf{0.761} &\textbf{0.767} &\textbf{0.771} &\textbf{0.774} &\textbf{0.777} &\textbf{0.780}& \textbf{0.742}\\
\bottomrule
\end{tabular}
\label{tab:tapos_val}
\end{table*}

\renewcommand{\arraystretch}{1.05}
\begin{table}[t]
\caption{Accuracy on the HMDB-51 and UCF-101 datasets.}
\centering
\setlength{\tabcolsep}{0.8mm}
\begin{tabular}{lcc}
\toprule
\multicolumn{1}{c}{} & \multicolumn{1}{c}{HMDB-51} & \multicolumn{1}{c}{UCF-101} \\ \midrule
\multicolumn{3}{l}{\textbf{Decoded video based methods} \textbf{\textit{(RGB only)}}} \\
ResNet-50 \cite{DBLP:conf/cvpr/resnet} & 48.9 & 82.3 \\
ResNet-152 \cite{DBLP:conf/cvpr/resnet} & 46.7 & 83.4 \\
ActionFlowNet (2-frames) \cite{DBLP:conf/wacv/NgCND18} & 42.6 & 71.0 \\
ActionFlowNet \cite{DBLP:conf/wacv/NgCND18} & 56.4 & 83.9 \\ 
PWC-Net + CoViAR \cite{DBLP:conf/cvpr/SunY0K18} & 62.2 & 90.6 \\
TVNet \cite{DBLP:conf/cvpr/FanHGEGH18} & 71.0 & 94.5 \\
C3D \cite{DBLP:conf/iccv/TranBFTP15} & 51.6 & 82.3 \\
Res3D \cite{DBLP:journals/corr/abs-1708-05038} & 54.9 & 85.8 \\
ARTNet \cite{DBLP:conf/cvpr/WangL0G18} & 70.9 & 94.3 \\ 
MF-Net \cite{DBLP:conf/eccv/ChenKLYF18} & 74.6 & 96.0 \\
S3D \cite{DBLP:journals/corr/abs-1712-04851} & 75.9 & 96.8 \\ 
I3D RGB \cite{DBLP:conf/cvpr/CarreiraZ17} & 74.8 & 95.6 \\
\midrule
\multicolumn{3}{l}{\textbf{Compressed video based methods}} \\
EMV-CNN \cite{DBLP:conf/cvpr/ZhangWW0W16} & 51.2 (split1) & 86.4 \\
DTMV-CNN \cite{DBLP:journals/tip/ZhangWWQW18} & 55.3 & 87.5 \\
CoViAR \cite{DBLP:conf/cvpr/coviar} & 59.1 & 90.4 \\
DMC-Net(ResNet-18) \cite{DBLP:conf/cvpr/dmc-net} & 62.8 & 90.9 \\ 
DMC-Net(I3D) \cite{DBLP:conf/cvpr/dmc-net} & 71.8 & 92.3 \\
E2E (ResNet-18) \cite{CVRL-GEBD-Li}   & 63.3 & 91.0 \\
E2E (I3D) \cite{CVRL-GEBD-Li}   & 72.1 & 92.5 \\
Ours (ResNet-18) & \textbf{63.8} & \textbf{91.4} \\
Ours (I3D) & \textbf{72.7} & \textbf{93.1} \\
\bottomrule
\end{tabular}
\label{table:results_of_action_recognition}
\end{table}

We conduct our experiments on the Kinetics-GEBD~\cite{DBLP:journals/corr/GEBD} and TAPOS~\cite{DBLP:conf/cvpr/TAPOS} datasets. The Kinetics-GEBD dataset contains the largest number of temporal boundaries, including $54,691$ videos and $1,290,000$ event boundaries, spans a broad spectrum of video domains in the wild and is open-vocabulary rather than building on a pre-defined taxonomy. The TAPOS dataset contains Olympics sport videos with 21 actions. The training set contains 13,094 action instances and the validation set contains 1, 790 action instances. Since it is not suitable for GEBD task, following \cite{DBLP:journals/corr/GEBD}, we re-purpose TAPOS for GEBD task by trimming each action instance with its action label hidden and conducting experiments on each action instance. Furthermore, to verify the generality and effectiveness of our method, we also conducted experiments on the popular action recognition datasets UCF101 \cite{DBLP:journals/corr/UCF101} and HMDB51 \cite{DBLP:conf/iccv/HMDB-dataset}. UCF101 consists of 101 action classes in $13,320$ videos, and HMDB51 contains $51$ distinct action categories with a total of $6,766$ video clips. 

To quantitatively evaluate the results of the generic event boundary detection task, the F1 score is used as the measurement metric. As described in \cite{DBLP:journals/corr/GEBD}, Rel.Dis. (Relative distance, the error between the detected and ground truth timestamps, divided by the length of the corresponding whole action instance) is used to determine whether a detection is correct (\ie, $\leq$ threshold) or incorrect (\ie, $>$ threshold). A detection result is compared with each rater's annotation, and the highest F1 score is treated as the final result. We report F1 scores of different thresholds range from 0.05 to 0.5 with a step of 0.05. In particular, all experimental results are the average of multiple experimental results.

\subsection{Implementation Detail}
We implement our method with the popular deep learning framework PyTorch~\cite{NEURIPS2019_9015_PYTORCH}. ResNet50 and ResNet18 \cite{DBLP:conf/cvpr/resnet} pretrained on ImageNet \cite{DBLP:conf/cvpr/imagenet} are used to extract the features for I-frames and P-frames in all experiments if not particularly indicated. Our method is implemented based on the MPEG-4 Part 2 specifications \cite{DBLP:journals/cacm/Gall91}, where each GOP contains $1$ I-frame and $11$ P-frames. We sample $3$ P-frames in each GOP to reduce the redundancy, \ie, $T=3$ in \eqref{equ:video_representation}. We use the standard SGD with momentum set to $0.9$, weight decay set to $10^{-4}$, and learning rate set to $10^{-2}$. We set the batch size to $4$ for each GPU and train the network on $8$ NVIDIA Tesla V100 GPUs, resulting in a total batch size of $32$. The network is trained for $30$ epochs with a learning rate drop by a factor of $10$ after $16$ epochs and $24$ epochs, respectively. We test the running speed of all methods on $1$ NVIDIA Tesla V100 GPU. All the source code of our method will be made publicly available after the paper is accepted.

We also evaluate our method with different backbones including CSN~\cite{DBLP:conf/iccv/TranWFT19} , ViT-Base~\cite{DBLP:conf/iclr/ViT}, and Swin-Tiny~\cite{swin} to compare with the state-of-the-art methods in the LOng-form VidEo Understanding Challenge (LOVEU).

\begin{figure*}[t]
\centering
\includegraphics[width=1.0\textwidth]{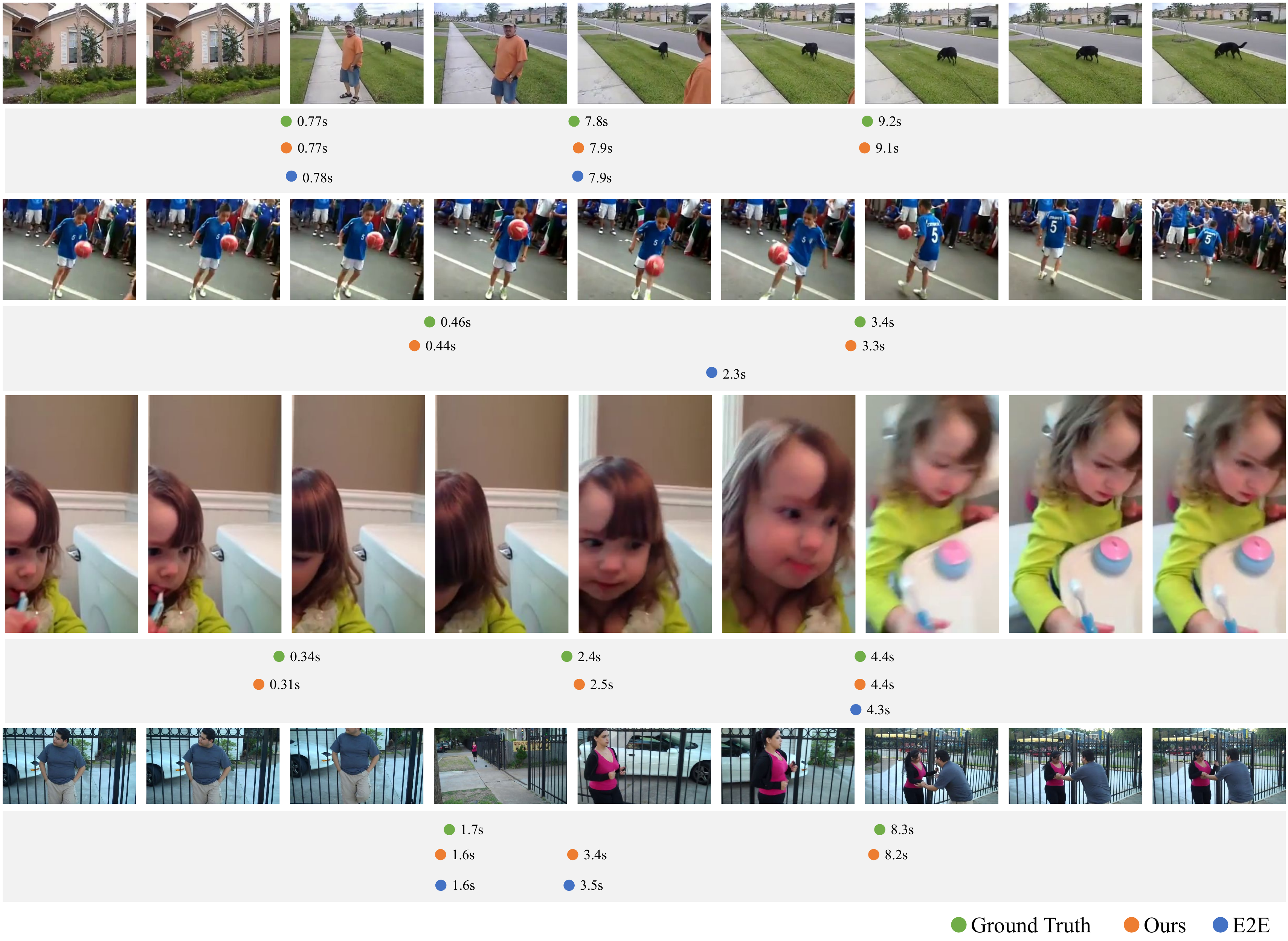}
\caption{Example qualitative results on Kinetics-GEBD validation split. Compared with E2E~\cite{CVRL-GEBD-Li}, our method can generate more accurate boundaries which are consistent with ground truth.}
\label{fig:results_visualization}
\end{figure*}

\subsection{Results and Analysis}
We first train and evaluate the proposed method on the Kinetics-GEBD \cite{DBLP:journals/corr/GEBD} train-validation split. The evaluation protocol presented in \cite{DBLP:journals/corr/GEBD} uses Relative Distance (\ie, \textbf{Rel.Dis.}, the error between the predicted and ground truth timestamps) to determine whether a prediction is correct or not and then use the precision, recall, and F1 scores as the evaluation metrics. We present all results with Rel.Dis. threshold set from 0.05 to 0.5 with 0.05 interval as shown in Table \ref{tab:results_of_kinetics_val}. Our method improves the F1 score over all thresholds by a large margin. Compared to the previous baseline method PC \cite{DBLP:journals/corr/GEBD}, our method achieves an absolute improvement of 14. 3\% while running $10\times$ faster. 

Compared to the previous end-to-end method E2E~\cite{CVRL-GEBD-Li}, our method achieves an absolute improvement of 2. 5\% while maintaining almost the same running speed. Improvements come mainly from the advanced spatial-channel compressed encoder and the local frames bag. Compared to the baseline method PC~\cite{DBLP:journals/corr/GEBD} with optical flow input stream, our advanced spatial-channel compressed encoder can be a better alternative to learning temporal information from cost-free motion vectors and residuals in compressed videos. Our local frames bag can also explicitly provide strong temporal signals for GEBD, giving 2. 1\% absolute improvements compared to E2E~\cite{CVRL-GEBD-Li}. Example qualitative results on Kinetics-GEBD are shown in Figure \ref{fig:results_visualization}. It's worth noting that our method has linear computational complexity with respect to the video length thus can scale well. We also evaluate our method on the TAPOS \cite{DBLP:conf/cvpr/TAPOS} train validation split. The results are shown in Table~\ref{tab:tapos_val}. Compared to DDM-Net~\cite{DBLP:journals/corr/progressive}, we increase the F1 score @ 0.05 from 0.604 to 0.618. Note that DDM-Net is not fully end-to-end and uses decoded RGB images as input, which is slow in both decoding and inference stages.

\begin{figure*}[t]
  \centering
   \includegraphics[width=1.0\linewidth]{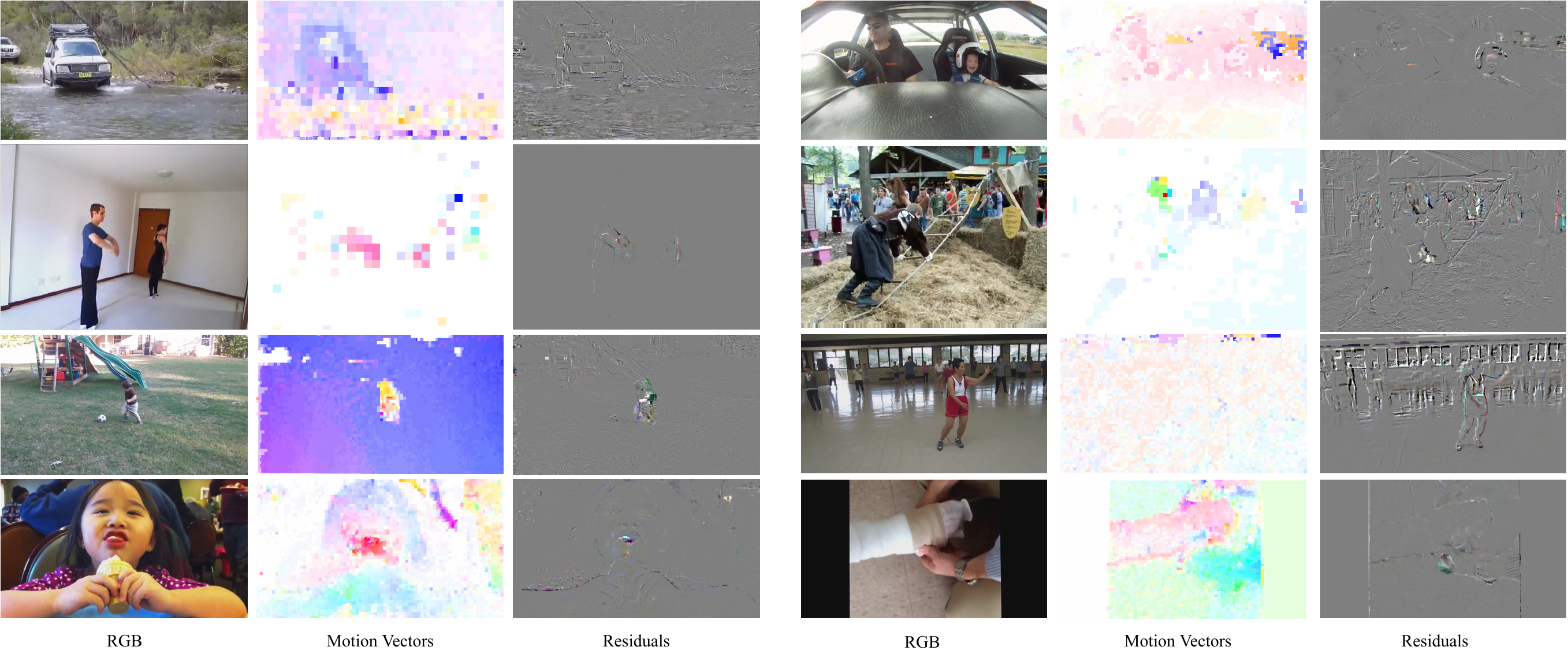}
   \caption{Visualization of the compressed information. The decoded RGB frames, motion vectors, and residuals are presented in different columns.}
   \label{fig:compressed_visualization}
\end{figure*}

We also conduct experiments on the UCF-101 and HMDB-51 action recognition datasets to validate the effectiveness of our method as in \cite{CVRL-GEBD-Li}. We follow the same settings as E2E \cite{CVRL-GEBD-Li} except that we use spatial-channel attention module (SCAM) to process the motion vectors and residuals instead of spatial-channel compressed encoder (SCCE). Our spatial-channel attention module gives more shortcuts to learning refined P-frame features from I-frame with motion vectors and residuals as guidance. Note that our local frames bag is also designed to capture temporal dependency, which is more suitable for event boundary detection. Thus, it is not applied in the action recognition task. We have two configurations for the action recognition task experiments, one with the same backbone as the event boundary, and the other replaced ResNet-18 with I3D to extract features from motion vectors and residuals, as shown in Table \ref{table:results_of_action_recognition}. Our method achieves competitive results compared to state-of-the-art methods in compressed domain, \ie, EMV-CNN \cite{DBLP:conf/cvpr/ZhangWW0W16}, DTMV-CNN \cite{DBLP:journals/tip/ZhangWWQW18}, CoViAR \cite{DBLP:conf/cvpr/coviar} and DMC-Net \cite{DBLP:conf/cvpr/dmc-net}. Compared to E2E \cite{CVRL-GEBD-Li}, we obtain about \textbf{0.5\%} improvements on both UCF-101 and HMDB-51 datasets with different backbones. The design principle of spatial-channel attention module (SCAM) is similar to spatial-channel compressed encoder (SCCE) \cite{CVRL-GEBD-Li} in that making model generate more discriminative P-frame representations with the guidance of the compressed information (motion vectors and residuals). However, motion vectors and residuals in compressed domain can be very noisy and it is tough to learn a beneficial representation for P-frames, as presented in Figure \ref{fig:compressed_visualization}. Our SCAM uses a gating mechanism to filter out noisy information in both spatial and channel dimensions and uses a bidirectional information flow to refine the origin P-frame feature. In this way, the noisy information from the features of I-frame, motion vectors, and residuals could be effectively and selectively fused together to generate high-quality P-frame features with little overhead. Compared to DMC-Net \cite{DBLP:conf/cvpr/dmc-net}, our method can directly learn discriminative features for P-frame with the spatial-channel attention module, which avoids extra optical flow as supervision during training phase.

\subsection{Failure Case Analysis}
\begin{figure*}[t]
\centering
\includegraphics[trim=60 100 688 150, clip, width=0.95\linewidth]{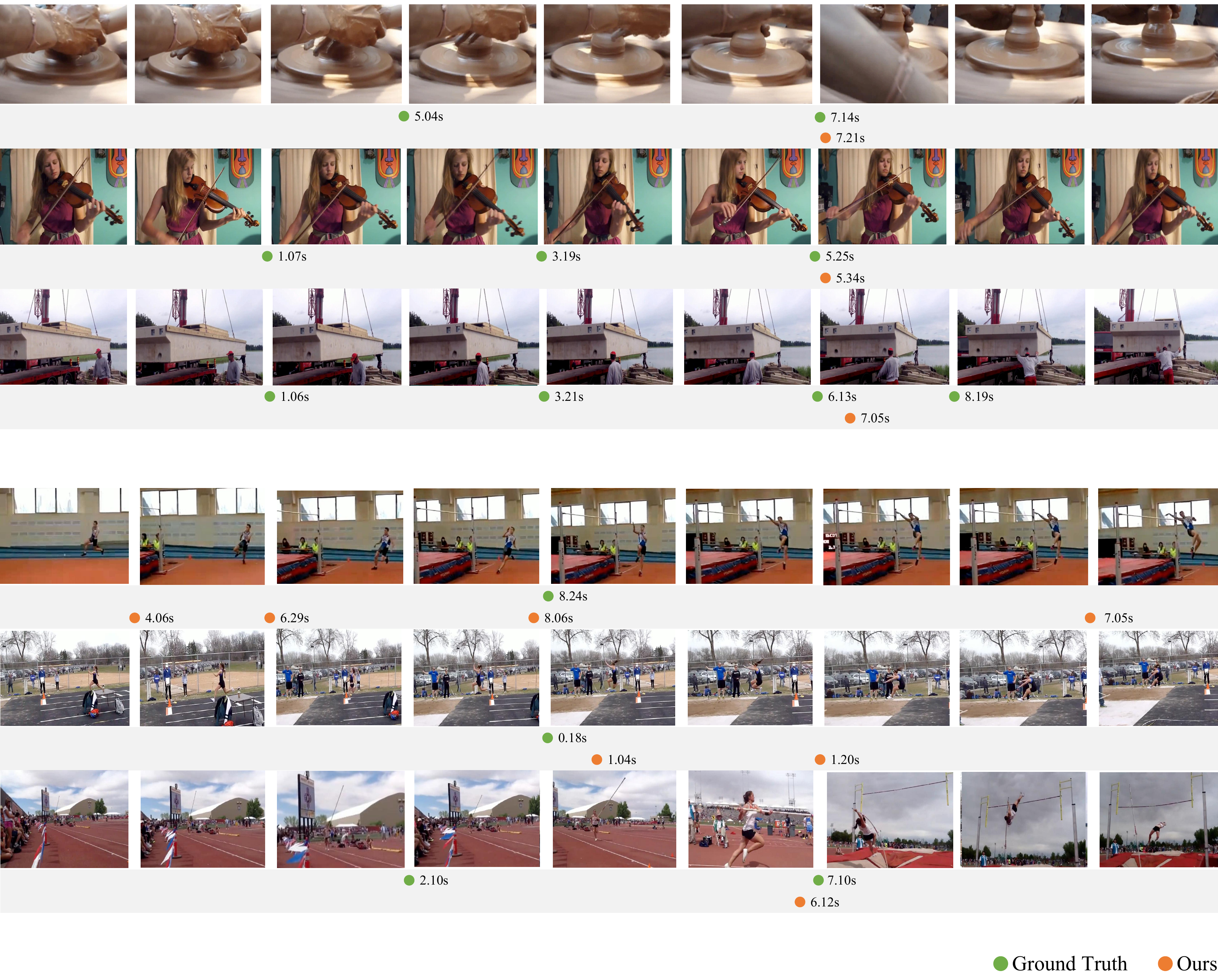}
\caption{Examples of failure cases taken from the validation split of the Kinetics-GEBD dataset.}
\label{fig:gebd_failed_example}
\end{figure*}

Based on our analysis of the Kinetics-GEBD validation split, we have identified two specific categories where our method tends to make mistakes. 
The first category includes videos with minimal visual changes, as shown in Figure \ref{fig:gebd_failed_example} (above): mold ceramics with clay, play the violin, and rotate a suspended object in one direction. These videos pose a challenge for our compression-based method as it struggles to detect event boundaries due to the loss of detailed information in the compressed domain. The second category comprises sports-related activity videos, exemplified by high jump, long jump, and pole vault in Figure \ref{fig:gebd_failed_example} (below). These movements typically involve three events: approach run, takeoff, and landing. However, our model tends to segment these actions into finer stages, such as dividing the action change in the event of takeoff into multiple events.
After thinking and research, we believe that the first type of problems can make up for the missing information by introducing new features such as optical flow. The second type of problem can be solved by introducing external common sense knowledge. For example, before performing event detection on high jump-related videos, let the model learn the knowledge that ``high jump is generally divided into three steps: approach run, takeoff and landing".

\subsection{Ablation Study}
In this section, we conduct several ablation studies to demonstrate the effectiveness of different components in the proposed method. All experiments are conducted on the Kinetics-GEBD train split with ResNet50 backbone and tested on a local minval split to reduce the computation cost. The local minval split is constructed from the Kinetics-GEBD validation split by randomly sampling $2,000$ videos. Default settings are marked in \colorbox{gray!20}{gray}.

\begin{figure}[t]
  \centering
   \includegraphics[width=0.9\linewidth]{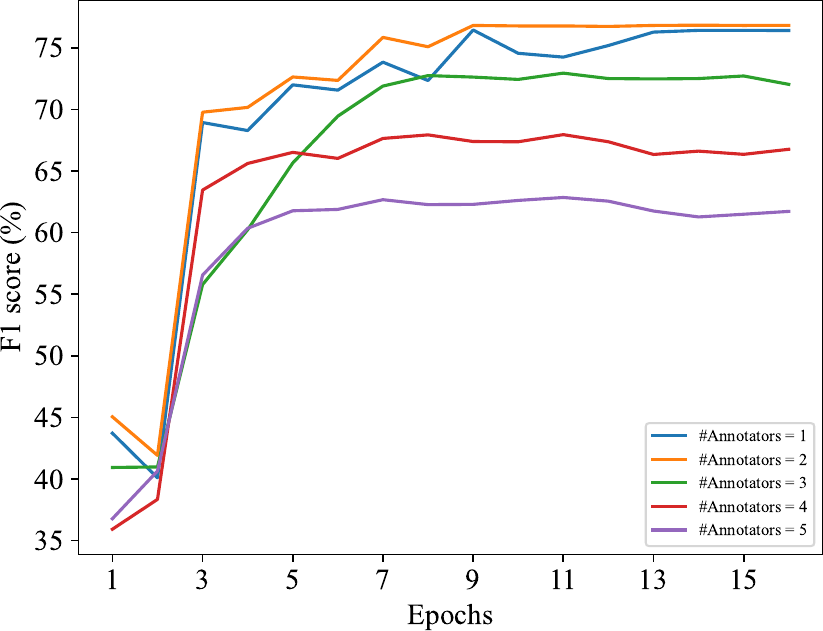}
   \caption{Influence of number of annotators. F1 scores get increased when using top-2 ground truth labels compared to only using top-1 labels. When using annotations from more than 3 annotators, the F1 score is decreased.}
   \label{fig:num_annotators}
\end{figure}

\textit{1) Influence of Number of Annotators:} To remedy the ambiguities of the event boundaries based on human perception, five different annotators are used for each video to label the boundaries based on predefined principles. To analyze the influence of number of annotators used in training phrase, we conduct 5 experiments by selecting top-1 to top-5 ground truth labels with respect to F1 consistency for training. The results are presented in Figure~\ref{fig:num_annotators}. We can see that the F1 scores increase when using top-2 ground truth labels compared to only using top-1 labels. This is because top-2 ground truth labels introduce more training samples and the annotation qualities are often consistent with each other. However, when using annotations from more than three annotators, the F1 score is decreased. This can be interpreted as different annotators have labeled event boundaries very differently based on their own subjectivity, which makes model confused and hard to converge. We utilize top-2 ground truth labels as the default setting in our experiments.

\begin{table}[t]
\caption{Influence of the backbone for compressed information.}
\centering
\setlength{\tabcolsep}{5.0pt}
\begin{tabular}{l|cccccc}
\toprule
Backbone &Rec &Prec &F1 & Speed(ms)\\
\midrule
\colorbox{gray!20}{ResNet-18}  &0.799 &0.740 &0.768 &4.5 \\
 ResNet-34  &0.797 &0.738 &0.766 &7.2 \\
 ResNet-50  &0.783 &0.725 &0.753 &12.1 \\
 ResNet-101  &0.781 &0.722 &0.750&18.3 \\
\bottomrule
\end{tabular}
\label{tab:compressed_backbones}
\end{table}

\textit{2) Influence of the Backbone for Compressed Information:} A intuitive benefit to use compressed information (\ie, motion vectors and residuals) is that we can learn discriminative features with a very lightweight backbone (\eg, ResNet-18), which saving model forwarding time naturally. However, \textit{can we further improve the F1 score if we use a more complex backbone for compressed information without considering time cost?} To explore this, we replace residuals and motion vectors' feature extractors $f_R$ and $f_M$ in section~\ref{sec:deformable_updating} with different backbones. The results are presented in Table~\ref{tab:compressed_backbones}. Interestingly, we do not observe obvious improvements when using much more power feature extractors and the performances even get slightly decreased when using ResNet-50 and ResNet-101. We attribute this to that motion vectors and residuals are noisy in nature and contains less useful information compared with fully decoded RGB frame, thus a more powerful backbone may overfit simple data distribution and cannot generalize well. We use ResNet-18 as default compressed information backbone for better speed and accuracy trade-off.

\begin{table}[t]
\caption{Influence of the temporal feature extractor.}
\centering
\setlength{\tabcolsep}{8.0pt}
\begin{tabular}{l|ccccc}
\toprule
Method &Rec &Prec &F1 \\
\midrule
 Average Pooling  &0.728 &0.672 &0.699 \\
 Max Pooling  &0.720 &0.669 &0.694 \\
 \midrule
$\varnothing$ \  &0.754 &0.731 &0.742 \\
 \colorbox{gray!20}{LSTM}  &0.799 &0.740 &0.768 \\
 GRU  &0.790 &0.732 &0.760 \\
 Transformer  &0.778 &0.725 &0.751 \\
\bottomrule
\end{tabular}
\label{tab:ablation_temporal}
\end{table}

\textit{3) Influence of the Temporal Feature Extractor:} After constructing local frames bag for each candidate frame as presented in section~\ref{sec:temporal_parser}, we are able to extract temporal context information using different modules. We analyze the effectiveness of various of temporal modules, the results are shown in Table~\ref{tab:ablation_temporal}. Average pooling and max pooling indicate that we accumulate local frames bag $\mathcal{B}^l$ into vector representation with $\operatorname{avg-pool}$ operator and $\operatorname{max-pool}$ operator, respectively. Thus the following group similarity module and FCN module are not adapted. Since $\operatorname{avg-pool}$ operator and $\operatorname{max-pool}$ operator only aggregate information linearly, which only can capture little temporal context, we observe obvious performance drop. $\varnothing$ indicates that the features of local frames bag $\mathcal{B}^l$ are directly fed into following group similarity module and FCN module, which cannot explicitly learn temporal context information, either. As for LSTM and GRU, we observe similar performance. When using more powerful module Transformer~\cite{DBLP:conf/nips/attisallyouneed}, we find that the F1 score is decreased. We infer that the Transformer may learn deleterious cues for generic event boundary detection. We use LSTM module if not specified. 

\begin{table}[t]
\caption{Influence of the local frames bag size $k$.}
\centering
\setlength{\tabcolsep}{10.0pt}
\begin{tabular}{c|ccccc}
\toprule
Bag size &Rec &Prec &F1 \\
\midrule
 $k=0$  &0.723 &0.650 &0.685 \\
 $k=2$  &0.734 &0.748 &0.741 \\
 $k=4$  &0.752 &0.749 &0.750 \\
 $k=6$  &0.775 &0.751 &0.763 \\
 \colorbox{gray!20}{$k=8$}  &0.799 &0.740 &0.768 \\
 $k=10$ &0.743 &0.752 &0.747 \\
$k=12$  &0.732 &0.758 &0.745 \\
\bottomrule
\end{tabular}
\label{tab:ablation_local_frames_bag}
\end{table}

\textit{4) Influence of the Local Frames Bag Size:} Besides the discriminative features of P-frames, the temporal dependencies are also important to predict the accurate event boundaries. To validate the effectiveness of the temporal contrastive module, we conduct several experiments, shown in Table \ref{tab:ablation_local_frames_bag}. As shown in Table \ref{tab:ablation_local_frames_bag}, without the temporal contrastive module (\ie, $k=0$), the overall accuracy (F1 score) decreased dramatically. After adapting the proposed temporal module, the F1 score improves sharply, \ie, $0.741$ {\it vs.} $0.685$ at $k=2$. To further analyze the effective of different window size in model accuracy, we also perform several experiments with different $k$ values. Table \ref{tab:ablation_local_frames_bag} shows that the recall starts to drop when $k>8$. We believe that it is because larger window size mixes temporal information cross boundaries, resulting in the combination of multiple different predictions and decreasing the recall value. Considering the performance, we set $k=8$ in our experiments as the default setting.

\begin{table}[t]
\caption{Effectiveness of different designs.}
\centering
\setlength{\tabcolsep}{3.0pt}
\begin{tabular}{l|cccc}
\toprule
Component & & & & \\
\midrule
 with SCAM?              &       &\ding{52}  &\ding{52} &\ding{52}  \\
 with Local Frames Bag?  &       & &\ding{52} &\ding{52} \\
 with Gaussian smoothing?  &       & &  &\ding{52} \\
 \midrule
 F1 Score &0.653&0.685&0.760&0.768 \\
\bottomrule
\end{tabular}
\label{tab:ablation_component}
\end{table}

\textit{5) Effectiveness of Various Model Design:} As discussed before, the Spatial Channel Attention Module (SCAM) generates discriminative features of P-frames with the guidance of motion vectors and residuals, which is critical for learning a good representation from compressed information. As shown in Table~\ref{tab:ablation_component}, the significant improvement in F1 score is obtained by using 
local frames bag, \ie, 0.760 vs. 0.685. The constructed local frames bag in Section~\ref{sec:temporal_parser} provides rich context information for event boundary detection. Compared to E2E~\cite{CVRL-GEBD-Li} that only uses linear weighted summation representation, our local frame bag is more flexible and adaptive for learning discriminative features leveraging group similarity and FCN. As shown in Table~\ref{tab:ablation_component}, adapting SCAM gives 0.32 improvement. Besides, using the soft labels generated by Gaussian kernel provides further 0.8\% absolute improvements. Using the ambiguous ``hard labels'' disrupt the learning process, which leads to poor convergence. Our soft label strategy effectively solve this issue and speeds up the training process.

\textit{6) Rationality of SCAM module design:} The spatial-channel attention module (SCAM) is designed to refine the feature representations of the P-frames based on compressed information with bidirectional information flow. From the Figure \ref{fig:scce}, the SCAM is composed of four modules: Guidance Information Module (GIM), Channel Updating Module (CUM), Spatial Updating Module (SUM), and Weight Updating Module (WUM). GIM processes input information, CUM and SUM introduce channel and spatial features, respectively, while WUM handles feature weighting. To verify the importance of these modules in SCAM, we conduct ablation experiments from two aspects of different inputs and different submodules, as shown in Table \ref{tab:scam_ablation}. From the Table \ref{tab:scam_ablation} (a),  we can find that inputs I-frame feature ${\bf x}_I$, motion vector feature ${\bf x}_{M}^t$ and resized motion vectors $M^t$ together work best. The results in Table \ref{tab:scam_ablation} (b) also prove the rationality and effectiveness of the design of each module in SCAM.

\begin{table}[h]\scriptsize
\caption{The ablation studies of SCAM on the Kinetics-GEBD minval split.}
\begin{subtable}[h]{0.5\textwidth}
\centering
\caption{Ablation of different inputs.}
\begin{tabular}{ccc|ccc}
\toprule
    $x_I$ & $M^t$ & $x_M^t$ & Rec & Prec & F1 \\ \hline
    \checkmark & - & -  & 0.795 & 0.722 & 0.757 \\
    - & \checkmark & - & 0.787 & 0.728 & 0.756  \\
    - & - & \checkmark & 0.795  & 0.717 & 0.754  \\
    \checkmark & - & \checkmark & 0.781 & 0.73 & 0.755  \\
    - & \checkmark & \checkmark & 0.764 & 0.739 & 0.752  \\
    \checkmark & \checkmark & - & 0.781 & 0.729 & 0.754  \\ 
    \checkmark & \checkmark & \checkmark & \bf{0.799} & \bf{0.740} & \bf{0.768} \\
\toprule
\end{tabular}
\label{tab:ab1}
\end{subtable}
\hfill
\begin{subtable}[h]{0.5\textwidth}
\centering
\caption{Ablation of different submodules.}
\begin{tabular}{ccc|ccc}
\toprule
    CUM & SUM & WUM & Rec & Prec & F1 \\ \hline
    \checkmark & - & - & \bf{0.809} & 0.664 & 0.729 \\
    - & \checkmark & - & 0.796  & 0.721 & 0.757  \\
    - & - & \checkmark & 0.743 & 0.354 & 0.480  \\
    \checkmark & - & \checkmark & 0.808  & 0.678 & 0.737  \\
    - & \checkmark & \checkmark & 0.789  & 0.729 & 0.758  \\
    \checkmark & \checkmark & - & 0.787 & 0.731 & 0.758  \\
    \checkmark & \checkmark & \checkmark & 0.799 & \bf{0.740} & \bf{0.768}\\
\toprule
\end{tabular}
\label{tab:ab2}
\end{subtable}
\label{tab:scam_ablation}
\end{table}

\textit{7) Influence of the fusing different modalities:}
Compressed video contains three kinds of I-frame feature ${\bf x}_I$, motion vectors $M^t$ and residuals $R^t$. In order to explore the effect of fusing different compressed domain information on model performance, we conducted three sets of comparative experiments and each consisting of two experiments: one with fusion and one without fusion. The experimental results in the Table \ref{tab:ablation_fuse} show that modality fusion can further improve the performance of the model. Specifically, fusing can bring about a $0.64$-$0.12$ improvement. This proves that fusing information from different compressed domains is beneficial for reconstructing the P-frame.

\begin{table}[ht]\scriptsize
\centering
\caption{Ablation of fusing different modalities on the Kinetics-GEBD minval split.}
\begin{tabular}{ccc|c|ccc} \toprule
${\bf x}_I$ & $M^t$ & $R^t$ & Fuse & Rec & Prec & F1 \\ \hline
\checkmark & \checkmark &  &  & 0.709 & 0.525 & 0.603  \\
\checkmark & \checkmark &  & \checkmark & 0.682 & 0.651 & 0.667  \\ \hline
 \checkmark & & \checkmark &  & 0.809 & 0.558 & 0.660  \\ 
 \checkmark & & \checkmark & \checkmark & 0.801 & 0.711 & 0.753  \\ \hline
\checkmark & \checkmark & \checkmark & & 0.753 & 0.572 & 0.650 \\
\checkmark & \checkmark & \checkmark & \checkmark & 0.799 & 0.740 & 0.768 \\
\toprule
\end{tabular}
\label{tab:ablation_fuse}
\end{table}

\textit{8) Influence of the sampling different P-frames in each GOP:}
In the MPEG-4 encoding format, each GOP usually contains 1 I-frame and 11 P-frames. If all P-frames are used as input, it will increase the amount of computation and slow down the inference speed. Therefore, we adopted the strategy of sampling part of the P-frames. However, too few P-frames will lose a lot of information and affect the performance of the model. In order to determine the optimal number of P-frames, we conducted experiments as shown in Table \ref{tab:ablation_T}. Specifically, we experimented with sampling 1 frame to 5 frames respectively, and the experimental results show that sampling 3 frames of P-frames works best. In addition, the result of sampling 5 frames is slightly reduced, $0.761$ VS $0.768$, which proves that too many P-frames will only bring redundant information and cannot further improve model performance.

\begin{table}[ht]\scriptsize
\centering
\caption{Ablation of sampling different P-frames in each GOP on the Kinetics-GEBD minval split.}
\begin{tabular}{c|ccc}
\toprule
T & Rec & Prec & F1 \\ \hline
1 & 0.749 & \bf{0.759} & 0.754 \\
2 & 0.800 & 0.711  & 0.753 \\
3 & 0.799 & 0.740 & \bf{0.768}  \\
4 & \bf{0.801} & 0.722 & 0.760 \\
5 & 0.797 & 0.728 & 0.761 \\
\toprule
\end{tabular}
\label{tab:ablation_T}
\end{table}

\begin{table}[t]
\caption{Comparisons with other state-of-the-art methods on LOVEU Challenge. $\dagger$ indicates that the results come
from our implementations since the test server is unavailable now. The speed (ms) is computed by averaging per-frame decoding and inference time.}
\centering
\setlength{\tabcolsep}{3.9pt}
\begin{tabular}{l|cccccc}
\toprule
Method &Rec &Prec &F1 & Speed\\
\midrule
$\dagger$CLA \cite{DBLP:journals/corr/abs-2106-11549} &0.815 &0.768 &0.791& 90.2 \\
$\dagger$CASTANET \cite{DBLP:journals/corr/abs-2107-00239} &0.838 &0.732 &0.781& 93.9 \\
 E2E (CSN+R18)~\cite{CVRL-GEBD-Li}  &0.813 &0.761 &0.786 &20.4 \\
 E2E (R50+R18)~\cite{CVRL-GEBD-Li}  &0.751 &0.742 &0.746 &4.7 \\
 \midrule
 Ours (CSN+R18)  &0.831 &0.792 & 0.812 & 20.3\\
 Ours (R50+R18)  &0.799 &0.740 &0.768 & 4.7\\
 Ours (ViT-B+R18)  & 0.798 & 0.727 & 0.761 & 5.9\\
 Ours (Swin-T+R18)  & 0.783 & 0.729 & 0.755 & 5.1\\
\bottomrule
\end{tabular}
\label{tab:compare_sota}
\end{table}

\subsection{Comparisons with State-of-the-arts on LOVEU Challenge} We also compare the proposed method with the state-of-the-art methods at CVPR’21 \textbf{LO}ng-form \textbf{V}id\textbf{E}o \textbf{U}nderstanding (LOVEU) Challenge\footnote{https://sites.google.com/view/loveucvpr21.}  as shown in Table \ref{tab:compare_sota}. The winners' solutions are complicated and running slowly, \eg, CLA \cite{DBLP:journals/corr/abs-2106-11549} relies on pre-extracted features and uses global similarity matrix which cannot scale well, CASTANET \cite{DBLP:journals/corr/abs-2107-00239} is not fully end-to-end and introduces redundant computations between nearby frames. E2E~\cite{CVRL-GEBD-Li} remedies this by using motion vectors and residuals in the compressed domain and achieves competitive results while running \textbf{20$\times$} faster than CLA \cite{DBLP:journals/corr/abs-2106-11549}. In addition, our method obtains absolute improvements of 2. 3\% and 4. 2\% compared to the preliminary version E2E~\cite{CVRL-GEBD-Li} when using CSN \cite{DBLP:conf/iccv/TranWFT19} and ResNet50 as backbones, respectively, while running almost at the same speed.

\subsection{Limitation of the compressed videos}
Working with compressed videos has several shortcomings that need to be considered, including potential loss of information during compressing videos into compressed streams. Since compressed data is a condensed version of uncompressed data, some data may be lost or altered, leading to inaccurate results and compromising data quality. Additionally, compression artifacts present a challenge when dealing with compressed data, as certain elements of the original data may be removed or distorted, particularly in visual tasks like image recognition. Finally, processing compressed data is generally more complex than uncompressed data because it must first be decompressed before analysis or manipulation. This added step can increase processing time and require specialized tools and techniques for accurate results. In addition, the limitations of directly using compressed video compared to decoding RGB video frames is that existing deep learning-based models are specifically designed to handle RGB videos. Therefore, directly applying these architecture may lead to suboptimal performance. A future direction is to develop a series of tailored network architectures directly suitable for compressed vision tasks.

Despite these challenges, working with compressed data remains crucial for research and development in various applications.

\section{Conclusion and Future work}
\label{sec:conclusion}
In this work, we propose an end-to-end compressed video representation learning method for GEBD. Specifically, we convert the video input into successive frames and use the Gaussion kernel to preprocess the annotations. Meanwhile, we design a spatial-channel attention module (SCAM) to make full use of the motion vectors and residuals to learn discriminative feature representations for P-frames with bidirectional information flow. After that, we propose a temporal contrastive module that uses local frames bag as representation to model the temporal dependency between frames and generate accurate event boundaries with group similarity. Extensive experiments conducted on the Kinetics-GEBD and TAPOS datasets demonstrate that the proposed method performs favorably against the state-of-the-art methods.

While our method has shown promising results, there is still room for improvement. Currently, the model leverages only the high-level semantic information extracted by the backbone network. However, for event boundary detection tasks, the importance of low-level detail information cannot be overlooked. Thus, an improved approach could involve the backbone network initially extracting multiscale features. Subsequently, a multiscale feature fusion module could be used to process these multiscale features. Regarding future research on further improvements, there are three possible directions: (1) The first direction for future research involves expanding current SCAM modules to support additional encoding formats beyond the current MPEG-4 standard. While the current SCAM can only handle one encoding format, extending it to include a range of general encoding formats would greatly enhance its utility and practicality. (2) The second direction involves enhancing the temporal module to allow independent and flexible selection of temporal modules based on specific scenarios. Currently, the temporal module is limited in its ability to select the most appropriate temporal module for different scenarios. (3) The last possible direction is to incorporate more information, such as audio and knowledge graphs. Audio information can provide a new basis for judgment and can assist the model in determining event boundary points. The knowledge information contained in the knowledge graph can help the model understand events to better determine the beginning and end of the boundary.

\section{Data Availability Statement}
The data that support the findings of this study are openly available in ``GEBD'' at \url{https://github.com/StanLei52/GEBD}, which are included in this published article \cite{DBLP:journals/corr/GEBD}.

\vspace{1em}
\noindent
\textbf{Acknowledgement.} Libo Zhang was supported by the Key Research Program of Frontier Sciences, CAS, Grant No. ZDBS-LY-JSC038, High-end Research Institutions Innovation Special Funds introduced by  Zhongshan Science and Technology Bureau (No.2020AG011) and Youth Innovation Promotion Association, CAS (2020111). Heng Fan and his employer received no financial support for the research, authorship, and/or publication of this article.

\bibliographystyle{sn-basic}
\bibliography{sn-bibliography}

\end{document}